\documentclass[letterpaper, 10 pt, conference]{ieeeconf}  

\IEEEoverridecommandlockouts                              

\usepackage{graphics} 
\usepackage{epsfig} 
\usepackage{mathptmx} 
\usepackage{times} 
\usepackage{amsmath} 
\usepackage{amssymb}  
\usepackage{graphicx}
\usepackage{threeparttable}
\title{\LARGE \bf
Uncertainty-aware Perception Models for Off-road Autonomous \\Unmanned Ground Vehicles
}

\author{Zhaoyuan Yang$^1$, Yewteck Tan$^1$, Shiraj Sen$^2$, Johan Reimann$^1$, John Karigiannis$^1$, \\Mohammed Yousefhussien$^3$, Nurali Virani$^{1*}$
\thanks{*Corresponding Author: \small\texttt{nurali.virani@ge.com}}
\thanks{$^1$GE Research, $^2$Motional, $^3$GE Digital}%
\thanks{$^{2,3}$Shiraj and Mohammed contributed to this work while being employed at GE Research}%
\thanks{Research was sponsored by the Army Research Laboratory (ARL) and was accomplished under Cooperative Agreement Number W911NF-20-2-0109. The views and conclusions contained in this document are those of the authors and should not be interpreted as representing the official policies, either expressed or implied, of the ARL or the U.S. Government. The U.S. Government is authorized to reproduce and distribute reprints for Government purposes notwithstanding any copyright notation herein.}%
}

\begin{document}

\maketitle
\thispagestyle{empty}
\pagestyle{empty}

\begin{abstract}

Off-road autonomous unmanned ground vehicles (UGVs) are being developed for military and commercial use to deliver crucial supplies in remote locations, help with mapping and surveillance, and to assist war-fighters in contested environments. Due to complexity of the off-road environments and variability in terrain, lighting conditions, diurnal and seasonal changes, the models used to perceive the environment must handle a lot of input variability. Current datasets used to train perception models for off-road autonomous navigation lack of diversity in seasons, locations, semantic classes, as well as time of day. We test the hypothesis that model trained on a single dataset may not generalize to other off-road navigation datasets and new locations due to the input distribution drift. Additionally, we investigate how to combine multiple datasets to train a semantic segmentation-based environment perception model and we show that training the model to capture uncertainty could improve the model performance by a significant margin. We extend the Masksembles approach for uncertainty quantification to the semantic segmentation task and compare it with Monte Carlo Dropout and standard baselines. Finally, we test the approach against  data collected from a UGV platform in a new testing environment. We show that the developed perception model with uncertainty quantification can be feasibly deployed on an UGV to support online perception and navigation tasks.

\end{abstract}

\section{Introduction}

To effectively perform autonomous navigation in complex environments, such as unstructured off-road terrains, the ability to perceive and build effective representations of the environment is crucial. Once an effective representation of the environment has been constructed, the complexity associated with developing and deploying other autonomous navigation algorithms, such as obstacle avoidance and optimize trajectories, becomes much simpler.

However, in real world environments, current Deep Learning-based perception models can struggle to provide the necessary information to generate an accurate representation of the environment, due to complexities such as:
\begin{enumerate}
    \item \textbf{Label Noise:} When using annotated data from single or multiple datasets, variability between human annotators, ambiguities in the images themselves, and differences in label definitions leads to a label noise and inconsistencies.
    \item \textbf{Hardware Noise:} The training data is not captured using the exact same camera as the target system and consequently the variability in the underlying hardware can introduce unwanted anomalies, such as differing ability to adjust exposure settings or ability to attenuate vehicle induced vibrations leading to blurry images.
    \item \textbf{Out-of-distribution Samples:} Encountering environment conditions (such as fog) and novel classes (such as bridges) not experienced during training can cause the perception models to produce inaccurate outputs.
\end{enumerate}

\begin{figure}[t!]
    \vspace{3pt}
	\begin{center}
		\includegraphics[scale=0.4]{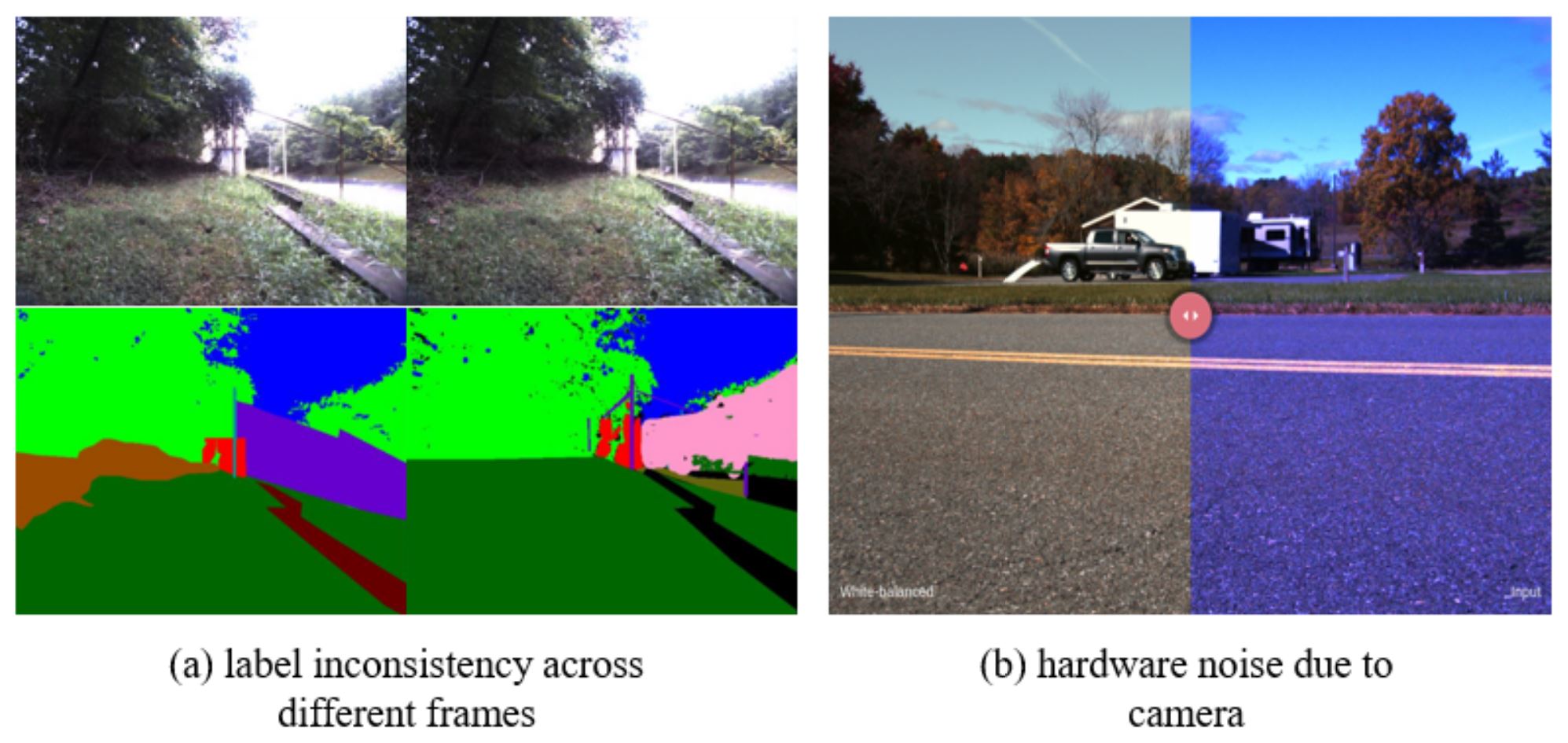}
	\end{center}
	\vspace{-10pt}
	\caption{Potential sources of noises for perception models - (a) shows an example of label inconsistency from a public dataset~\cite{RUGD2019IROS}, and (b) shows an example of color inconsistency due to camera parameters.}
	\label{fig:motivation}
	\vspace{-10pt}
\end{figure}

Figure \ref{fig:motivation} shows some noise/inconsistencies caused by labels and hardware settings. When constructing perception models, these factors must be considered and minimized. However, in some instances the introduced uncertainty is unavoidable and consequently the overall system must be designed to accommodate any potential adverse impact. 

To design a system that can accommodate the inherent environment variation and uncertainty, we present a methodology by which the epistemic uncertainty in the perception pipeline can be captured and used in real-time to improve autonomous navigation through unstructured environments. The system is trained without using a simulation environment or synthetic data, that is, the perception model is trained directly on several diverse datasets of captured images and human provided annotations. 

A key insight of the work is that perception models trained using a single dataset typically do not generalize well to other datasets and that directly training models by combining multiple datasets is problematic due to inconsistent labels and label noise across datasets.  Consequently, we show that training perception models to capture epistemic uncertainty in the output leads to improved model performance. To the best of our knowledge, the presented investigation into the impact of uncertainty quantification (UQ) when a model is trained on multiple datasets with different label sets has not been done in literature up to this point.

After briefly covering related work, we present a detailed description of the methodology used to capture the perception model uncertainty. Once the theoretical foundation has been established, we describe results on several off-road perception datasets, where we compare the segmentation performance using our uncertainty-aware perception pipeline against a baseline model. We conclude by providing some suggestions for further developments and experiments that could be performed along with concluding remarks.

\section{Background and Related Work}

\subsection{Datasets for Building Perception Models}
There has been some recent work in off-road autonomous navigation using machine learning-based perception models~\cite{tan2021risk,cai2022risk}. However, the efforts focused on off-road navigation are still less extensive compared to on-road, due to the lack of off-road consistent and structured datasets. The main datasets available for training a semantic segmentation model for off-road autonomous navigation are: RUGD \cite{RUGD2019IROS}, RELLIS-3D \cite{jiang2021rellis}, DeepScene-Freiburg-Forest \cite{valada16iser}, YCOR \cite{maturana2018real}, SOOR \cite{SOOR2021} and very recently ORFD \cite{ORFD2022}. 
These datasets demonstrate certain differences with respect to the number of classes present in the data as well as variation on image resolution and aspect ratio. In this work, we combine multiple datasets together to build more robust segmentation models. 
An extensive review of fusing multiple data sources for segmentation is presented in \cite{Feng2021}. In this work, we focus on semantic segmentation with RGB images only.

\subsection{Combining Multiple Datasets for Robust Model}
One approach to improve the model robustness is to combine multiple datasets for training. An interesting work presented in \cite{Egocentric2022} combines a broad set of datasets with over 40K RGB images. The datasets cover urban scenery \cite{Cityscapes2016, BDD2020, Mapillary2017, acdc2021}, off-road domains \cite{RUGD2019IROS, maturana2018real, TAS2021} as well as mixed scenery \cite{IDD2018}. Authors combine all these datasets that contain different clusters of classes by mapping the initial labels to only three classes, each represents a level of driveability (i.e. Preferable, Possible, Impossible). The reported results are promising but they lack finer resolution of labels that could enable reasoning for traversability (e.g., grass is easier to traverse for the robot compared to gravel). UQ was not considered in that work. 

In~\cite{Youngsaeng2021}, a semantic segmentation method is proposed with a modified encoder-decoder architecture. The architecture was trained with the RUGD dataset and the RELLIS-3D dataset. In~\cite{SOOR2021}, the authors combine DeepScene-Freiburg-Forest dataset and SOOR dataset for training. They fuse RGB and texture descriptors (i.e. local binary pattern descriptors)~\cite{Ojala2002}, which eliminates the requirement for geometric matching needed when using different sensor data inputs. They report improved segmentation performance due to the fusion compared to only RGB data segmentation. 

\subsection{Capturing Uncertainty in Perception Model}
Apart from combining multiple datasets, quantifying uncertainty in the model could also provide robustness to the model. Epistemic uncertainty and aleatoric uncertainty are two major types of uncertainty for perception models~\cite{kendall2017uncertainties}. Epistemic uncertainty captures the uncertainty in the model while aleatoric uncertainty captures the noise in the observations. Introducing more training data could reduce the epistemic uncertainty; however, aleatoric uncertainty will not be explained away by more training data. 

Monte Carlo (MC) Dropout~\cite{gal2016dropout} and Deep ensembles~\cite{lakshminarayanan2017simple} are two widely applied approaches to quantify the uncertainty. Both methods are first demonstrated on the classification task as well as the regression task then later extend to other tasks. Bayesian SegNet~\cite{BMVC2017_57} extends MC Dropout to segmentation tasks by inserting dropout sampling in feature layers of the segmentation models. Both MC Dropout and Deep ensembles aim to characterize the epistemic uncertainty of a model. Both approaches have their advantages and disadvantages. Deep ensembles generates deterministic results but it requires evaluating multiple models during inference which could be computationally expensive. MC Dropout only needs a single model but it requires sampling from a probability distribution which could lead to high variance in predictions. To balance the trade-off between computation and variance, the Masksembles approach has been recently proposed in literature~\cite{durasov2021masksembles}. Instead of sampling weights from a probability distribution like MC Dropout, Masksembles applies a set of fixed masks to model weights. 

\subsection{Incorporating Semantic Uncertainty in Planning}
Risk-aware navigation has been discussed in multiple works~\cite{xiao2019, pereira2013, chung2019}. We showed one way of using uncertainty from perception models for navigation planning in~\cite{tan2021risk}. An overview of that approach is included here for completeness. A synchronized image and point cloud is provided by sensors on the UGV. RGB data-based terrain segmentation models are used to segment the image into various terrain types that are informative to determine traversability for navigation. The label uncertainty is estimated for the terrain segmentation that provides the associated confusion among the various terrain types for each pixel. The output of the terrain segmentation is projected to multiple Cartesian maps by using the lidar and camera calibration. In parallel, geometry-based models are used to segment the point cloud to find traversable regions by binning the grids based on the height and slope of the terrain. The terrain maps from the camera and lidar are fused using a probabilistic framework that accumulates multiple sources of information over time for every grid cell to generate a cost map that takes the the segmentation uncertainty into account. The costmap is used by a risk-aware path planner that plans paths based on the associated risk and transitions to either exploratory behaviors (if no suitable paths are found) or SOS behavior (if the planner has run out of all options) to get help from a human teammate. This allows the robot to make progress and ask for help without jeopardizing the mission or increasing the cognitive load on the human teammates. In this paper, we are focusing only on training perception models and obtaining the label uncertainty for the semantic segmentation model. Details on the navigation planner, fusion with lidar, and intelligent behaviors is beyond the scope of this paper. Interested readers can check out~\cite{tan2021risk} for more details.

\section{Technical Approach}
\subsection{Epistemic Uncertainty Quantification (UQ)}
In this section, we describe the epistemic UQ methods. Bayesian method provide a mathematically grounded framework to quantify the uncertainty. We will briefly discuss the Bayesian models in the following section. We use bold upper-case letters to represent matrices and bold lower-case letters to represent vectors. Let $\mathbf{X}$ and $\mathbf{Y}$ represent the training data matrix and their corresponding training labels. During inference, given an input data vector $\mathbf{x}$, the Bayesian models aim to predict the posterior distribution of the probability vector $\mathbf{y}$, where each element of the vector $\mathbf{y}$ represents probability of an input vector $\mathbf{x}$ belongs to a specific class. The posterior distribution can be obtained as follows: 
\begin{align*}
	p(\mathbf{y}|\mathbf{x}, \mathbf{X}, \mathbf{Y}) = \int p(\mathbf{y}|\mathbf{x}, \mathbf{W}) p(\mathbf{W}|\mathbf{X}, \mathbf{Y}) d\mathbf{W}   \tag{1}
    \label{eq:1}
\end{align*}
where, $\mathbf{W}$ represents the model weights. In many situations, posterior distribution $p(\mathbf{W}| \mathbf{X}, \mathbf{Y})$ is not tractable; thus, approximations of the posterior distribution $p(\mathbf{y}|\mathbf{x}, \mathbf{X}, \mathbf{Y})$ are often obtained. In this work, we consider MC Dropout~\cite{gal2016dropout} and Masksembles~\cite{durasov2021masksembles} as methods to approximate the posterior distribution. Both approaches are efficient to deploy on platforms with limited computing resources. 

\noindent
\textbf{Monte Carlo (MC) Dropout:} Instead of calculating the integral over all possible weights $\mathbf{W}$ in the equation~\eqref{eq:1}, MC Dropout runs dropout operations multiple times to estimate the integral. Let $f(\cdot;\mathbf{W}_t)$ represents the function parameterized by a neural network with weight $\mathbf{W}_t$ which takes an input data vector $\mathbf{x}$ and output a probability vector $\mathbf{y}$. The posterior of the distribution $\mathbf{y}$ can be approximated as follows:
\begin{align*}
	p(\mathbf{y}|\mathbf{x}, \mathbf{X}, \mathbf{Y}) \approx \frac{1}{T} \sum_{t=1}^T f(\mathbf{\mathbf{x}; \mathbf{W}_t}), \tag{2}
	\label{eq:2}
\end{align*}
where, $T$ represents the number of samples and a sample of weight variable is obtained by generating a Boolean matrix $\mathbf{B}_t$ derived from sampling Bernoulli distribution and then computing element-wise product $\odot$, i.e., $\mathbf{W}_t=\mathbf{B}_t \odot\mathbf{W}$. The model weight $\mathbf{W}$ is obtained through standard model training with dropout regularization~\cite{gal2016dropout}.

\begin{figure}[h!]
	\vspace{-6pt}
	\begin{center}
		\includegraphics[scale=0.35]{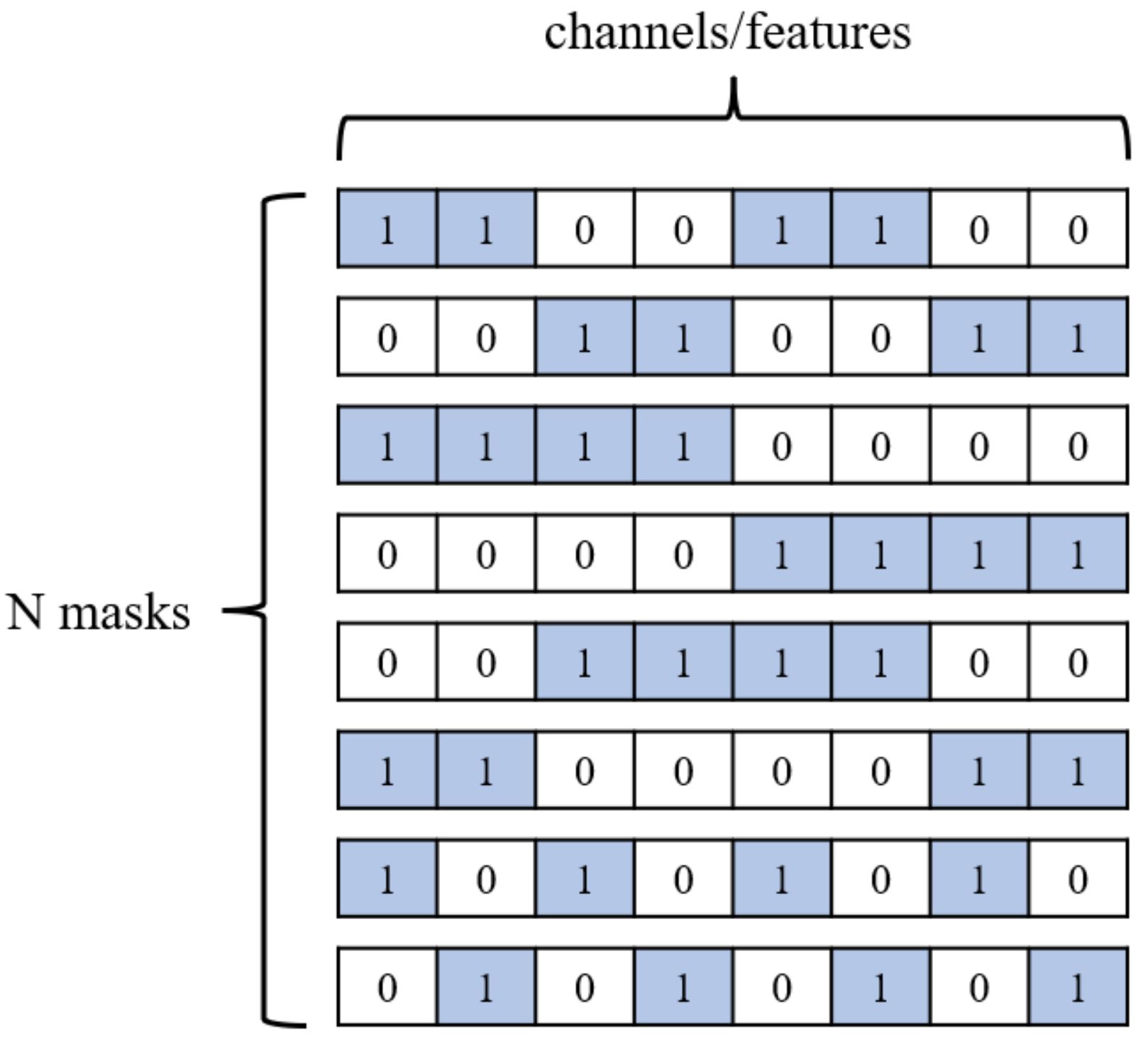}
	\end{center}
	\vspace{-6pt}
	\caption{Example of masks used in the Masksembles approach. Each row here represents a mask, 1 represents value is kept and 0 represents value is masked.}
	\label{fig:maskexample}
	\vspace{-6pt}
\end{figure}

\noindent
\textbf{Masksembles:} To obtain a stable estimation of the integral with low variance, MC Dropout needs large number of samples. However, it is infeasible to sample large number of weights during inference. Limited number of samples could lead to small diversity in weight samples; consequently, the prediction could be biased. To address this challenge, we study the Masksembles approach which uses a set of deterministic masks instead of stochastic masks to obtain the model weight samples for both training and testing \cite{durasov2021masksembles}. Masks for the Masksembles approach are designed to minimize the overlap. Figure \ref{fig:maskexample} shows a set of masks applied to mask out the channels of a convolutional neural network where overlap of masks are minimized. 

Assuming there are $C$ classes in total, and $y_i$ represents the probability of the input vector $\mathbf{x}$ belongs to class $i$. Once we obtain the probability vector $\mathbf{y} = [y_1, y_2, ..., y_C]^\intercal$, there are many measurements can be applied to estimate uncertainty. Common measurements include entropy, variation ratio, mean standard deviation, etc. In this work, we use entropy as our uncertainty measurement and uncertainty of the output vector can be estimated by $H(\mathbf{y}) = -\sum_{i=1}^C y_i \text{ log } y_i$.

\section{Experiments and Implementation}

In this section, we will cover the important implementation details and the key results of our work. First, details related to training data consolidation across the multiple datasets is described with emphasis on how the different labels are translated between the different collections. Once the data preprocessing has been covered, we describe how the training of the segmentation model is performed. Finally, we evaluate the trained models with multiple public datasets and with data collected using a UGV platform at a local testing site.

\subsection{Label Conversion across Different Datasets}
\begin{figure}[h!]
	\begin{center}
		\includegraphics[scale=0.46]{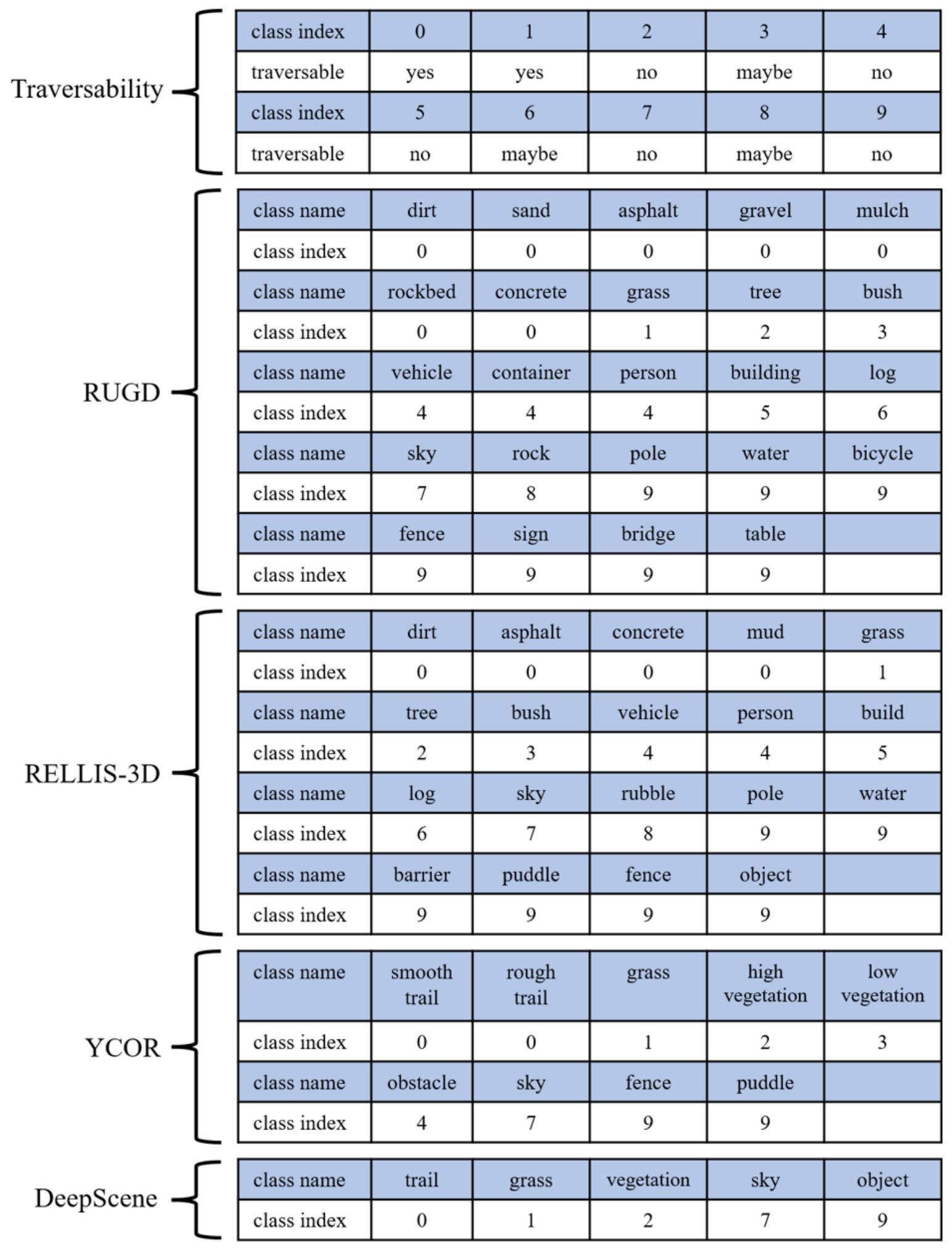}
	\end{center}
	\vspace{-6pt}
	\caption{Traversability of classes and label conversion from different datasets. Class name represents label names used in the original dataset and class index represents labels after conversion.}
	\vspace{-6pt}
	\label{fig:labelconv}
\end{figure}

\begin{figure}[h!]
	\vspace{-10pt}
	\begin{center}
		\includegraphics[scale=0.4]{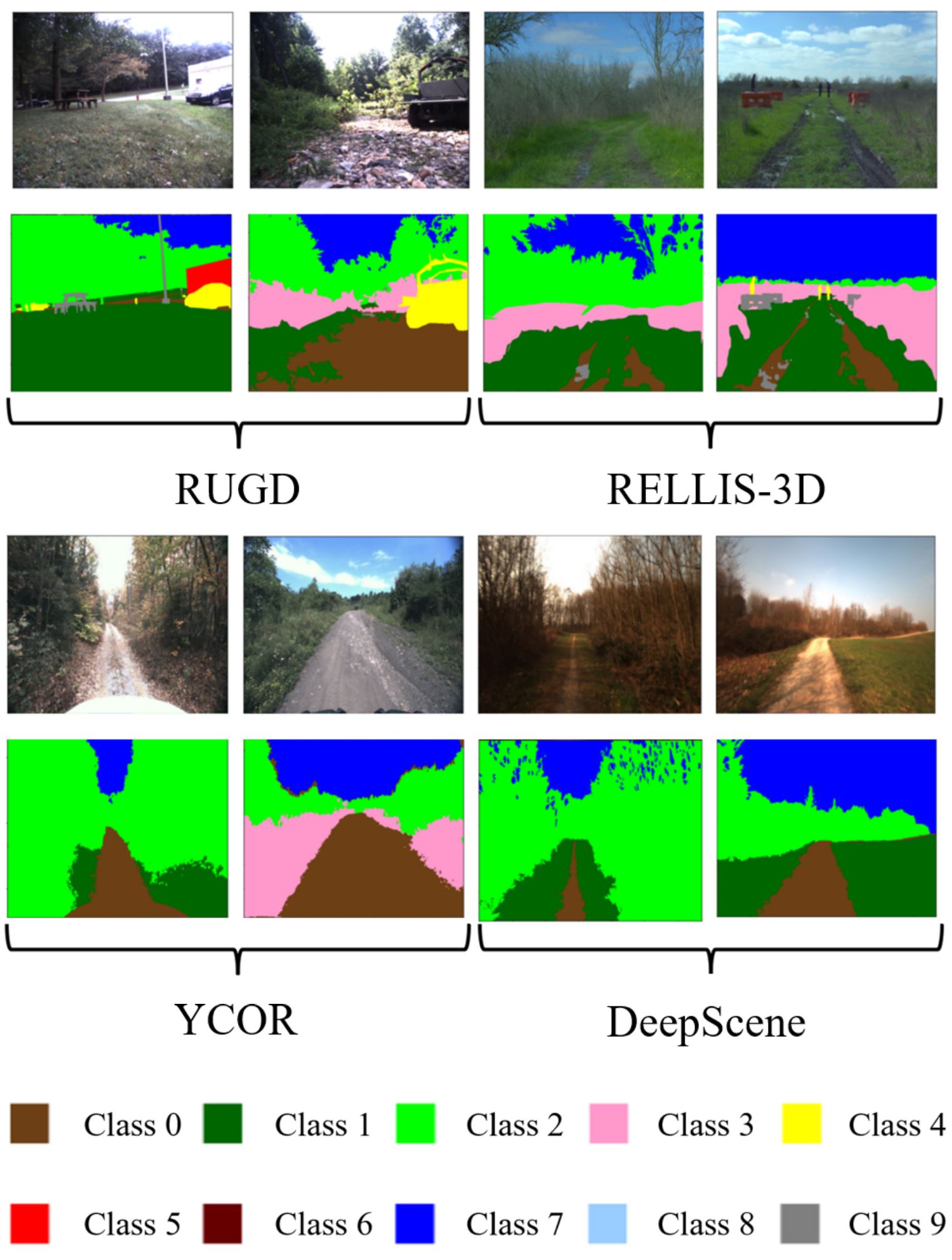}
	\end{center}
	\vspace{-10pt}
	\caption{Example images from different datasets and their converted labels.}
	\vspace{-10pt}
	\label{fig:sampledata}
\end{figure}

We use the following dataset for our experiments: RUGD \cite{RUGD2019IROS}, RELLIS-3D \cite{jiang2021rellis}, DeepScene \cite{valada16iser}, YCOR \cite{maturana2018real}. These off-road navigation datasets have different label classes and styles. For example, a region with bushes is labeled as ``bush" in  RUGD and  RELLIS-3D, but it is called ``low vegetation" in YCOR, whereas DeepScene labels both trees and bushes as ``vegetation". Therefore, to combine different datasets for training and evaluation, a label mapping must be established for each dataset. To address this problem, we define mappings based on appearances, traversability, and frequency of occurence. We show the traversability as well as mappings between different datasets in Figure \ref{fig:labelconv} and some examples as well as their labels in Figure \ref{fig:sampledata}. The mapping for class index to traversability is adapted based on Tan \textit{et al}~\cite{tan2021risk}, however other mappings could also be applied. 

\subsection{Semantic Segmentation and Uncertainty Quantification.}
To show impact of combining datasets for training, we train multiple models with a single dataset and multiple datasets. For models trained with a single dataset, only the RUGD dataset is used. For models trained with multiple datasets, all datasets are used.

To keep a reasonable resolution for navigation and make the model computational efficient on our platform, we set dimension of the input image be $672 \times 544$. We use random rotation, horizontal flip, random brightness and contrast as well as motion blur to augment our training data. 
We use U-Net architecture~\cite{ronneberger2015u} with the pre-trained EfficientNet-b3 model~\cite{tan2019efficientnet} as the backbone instead of other computationally expensive models as we need to deploy the model on a platform with limited computing resources. We modify the model~\cite{Yakubovskiy2019} and follow the method from Bayesian SegNet work~\cite{BMVC2017_57} to add the MC Dropout or Masksembles in the center/bottleneck layer of the U-Net. To fix the variance caused by differences in models initialization, all models trained in our experiments are initialized with the same weights and training configurations are the same for all experiments. For MC Dropout and Masksembles, half of the outputs are masked during training (dropout rate set to 0.5). We use SGD with momentum of 0.9 to optimize models. Objective function is chosen to be a sum of Jaccard loss and Focal loss (with $\gamma = 2$)\cite{lin2017focal}. We observe that this combined loss performs better for imbalanced datasets. At beginning of training, we freeze the pre-trained encoder (EfficientNet) and only train the decoder. We set batch size to be 6 and train the model for 160 epochs. The learning rate increases from $10^{-4}$ to $10^{-2}$ in the first 2 epochs, and we use the polynomial decay with power of 0.9 for the rest of training. Once training of the decoder is completed, we start to jointly train the encoder-decoder with the same training configuration. The optimal model is selected based on the mean intersection over union (mIOU) metric on validation data. We observe that learning rate scheduling as well as train the decoder before jointly train the encoder-decoder together provides a better performance than other training configurations. 

\begin{figure}[h!]
	\begin{center}
		\includegraphics[width=0.95\columnwidth]{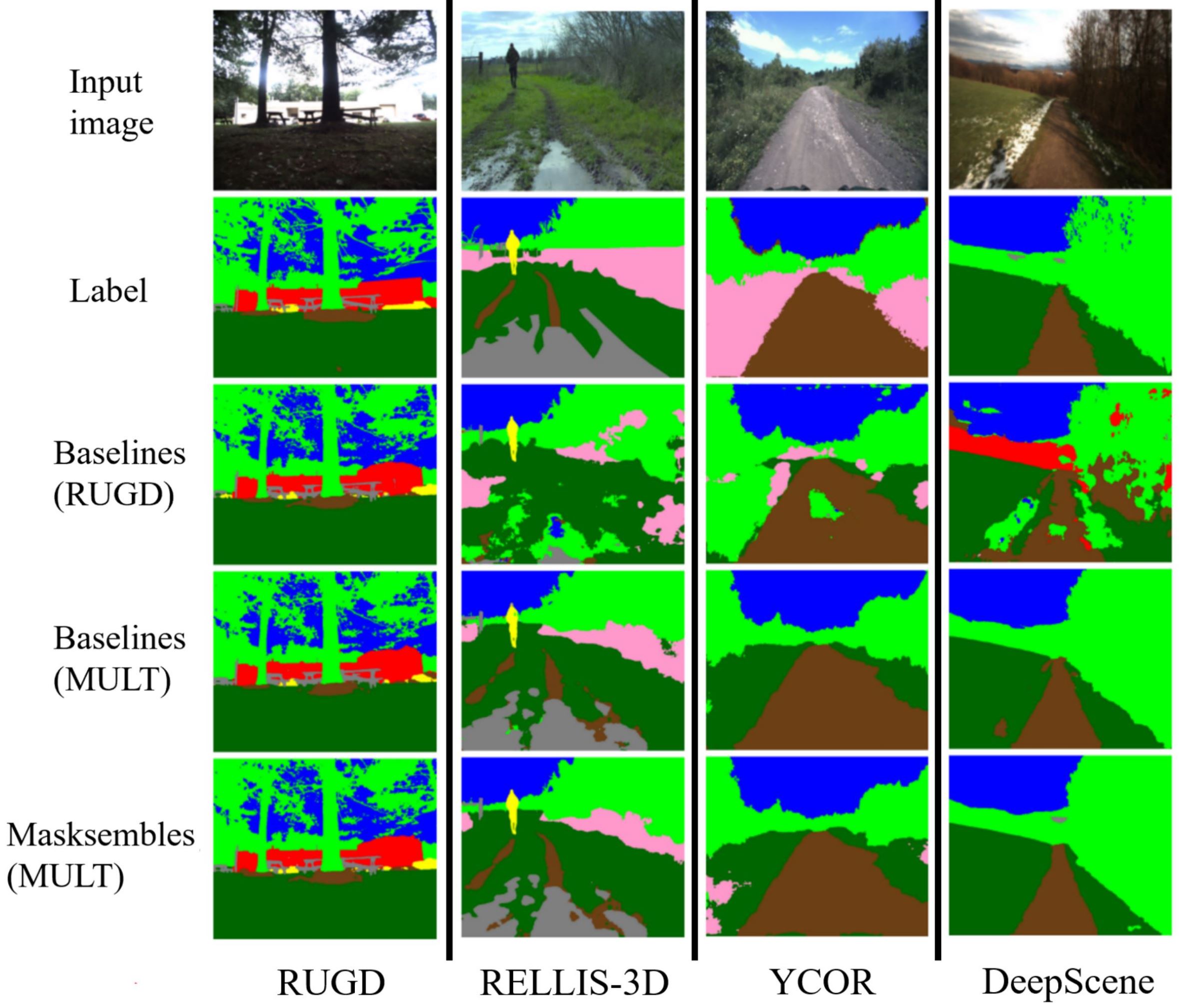}
	\end{center}
	\vspace{-6pt}
	\caption{Comparison of results from different models across multiple datasets. Each column contains segmentation predictions from different models for an image. First row contains input images, second row contains annotated labels, and third to fifth rows contain predictions from different models where (RUGD) represents models only trained on the RUGD dataset while (MULT) represents models trained on multiple datasets. }
	\label{fig:samplepred}
	\vspace{-6pt}
\end{figure}

 Figure~\ref{fig:samplepred} shows some example predictions of trained models: 1) standard model only trained on the RUGD dataset, 2) standard model trained on multiple datasets, and 3) Masksembles trained on multiple datasets. Since part of RUGD dataset is available for all models during training, we discover that all three models perform relatively well on the RUGD dataset. However, unlike standard model and the Masksembles model trained with multiple datasets, the standard model only trained with one dataset does not generalize well to the other datasets. One of reasons for this phenomenon is that the distribution of RUGD dataset based on time of day, locations, and seasons does not represent environments in the other datasets. Thus, combining multiple datasets should provide more diversity for training. 
 
\begin{figure*}[h!]
	\begin{center}
		\includegraphics[width=0.95\textwidth]{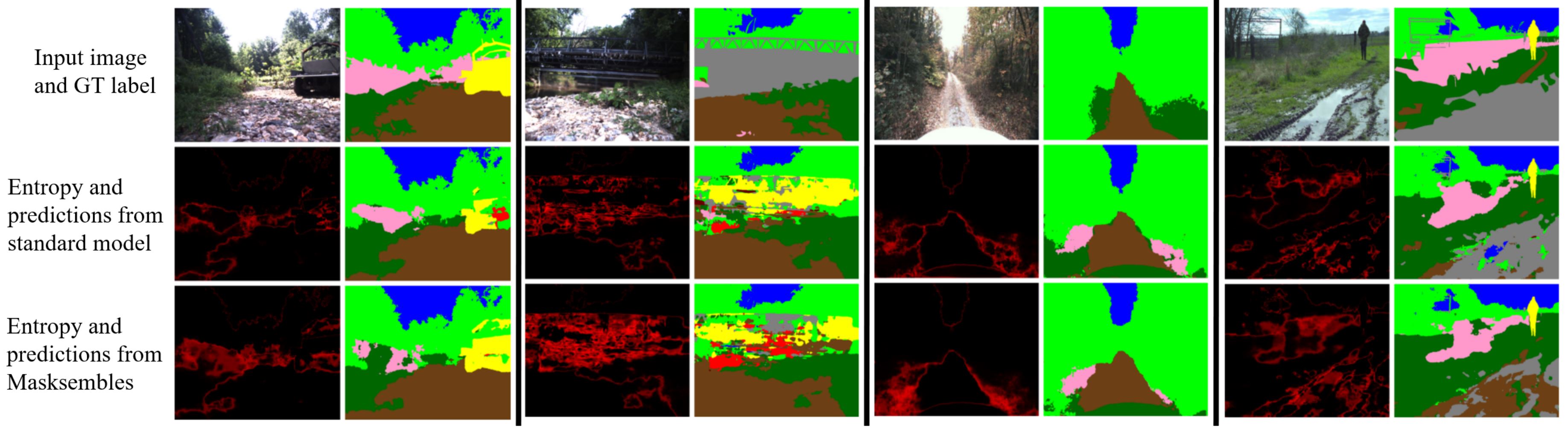} 
	\end{center}
	\vspace{-6pt}
	\caption{Uncertainty estimation of test data from public datasets. We visualize the entropy of the standard model and entropy of the Masksembles trained on multiple datasets. First row contains input images and ground truth labels. Second row contains entropy of predictions and prediction results from the standard model. Third row contains entropy of predictions and prediction results from the Masksembles. We use red color to represent uncertainty where higher intensity represents higher entropy. Note that the Masksembles has a higher entropy when the model makes an incorrect prediction.}
	\label{fig:uqvis}
	\vspace{-6pt}
\end{figure*}

In the following sections, we evaluate models on public datasets to show quantitative performance of different models. We demonstrate performance of segmentation models using global accuracy (G), class average accuracy (C) and mIOU. Baselines are standard models trained with a single dataset and multiple datasets. Uncertainty-aware models include MC Dropout and Masksembles trained on a single dataset and multiple datasets. Both MC Dropout and Masksembles run 8 sampling iterations during inference. We optimize models with TensorRT and inference speed is around 0.2s per image for MC Dropout and Masksembles on a Tesla P100 GPU. The performances on validation data are shown in Table \ref{tbl:valperformance} and performances on testing data are shown in Table \ref{tbl:testperformance}.

\begin{table}[h!]
	\centering
	\tabcolsep=3pt
	\caption{Performance of models on validation data.}
	\label{tbl:valperformance}
	\resizebox{\linewidth}{!}{
		\begin{tabular}{l|ccc|ccc|ccc}
			& \multicolumn{3}{c|}{Baselines} & \multicolumn{3}{c|}{MC Dropout} & \multicolumn{3}{c}{Masksembles} \\
			Dataset & G & C & mIOU & G & C & mIOU & G & C & mIOU\\
			\hline \hline
			Val RUGD (RUGD) &  92.33 &  69.73 &  60.89 &  92.31 &  \textbf{73.15} &  \textbf{61.80} &  \textbf{92.38} &  71.53 &  61.61 \\
			Val RUGD (MULT) &  92.13 &  70.22 &  60.04 &  92.22 &  71.22 &  61.23 &  92.35 &  70.52 &  61.11 \\
			\hline 
			Val RELLIS (RUGD)	 &  73.57 &  41.17 &  29.11 &  74.92 &  41.83 &  29.29 &  75.70 &  44.35 &  30.22 \\
			Val RELLIS (MULT) &  86.31 &  \textbf{72.91} &  \textbf{59.12} &  86.07 &  72.24 &  57.72 &  \textbf{87.65} &  71.66 &  57.57 \\
			\hline
		\end{tabular}
	}
	\begin{tablenotes}
	\item\textbf{} 
    \item\textbf{Table Notes: } (RUGD) represents models only trained on the RUGD \\dataset while (MULT) represents models trained on multiple datasets.
    \end{tablenotes}
	\vspace{-6pt}
\end{table}

\begin{table}[h!]
	\centering
	\tabcolsep=3pt
	\caption{Performance of models on testing data.}
	\label{tbl:testperformance}
	\resizebox{\linewidth}{!}{
		\begin{tabular}{l|ccc|ccc|ccc}
			& \multicolumn{3}{c|}{Baselines} & \multicolumn{3}{c|}{MC Dropout} & \multicolumn{3}{c}{Masksembles} \\
			Dataset & G & C & mIOU & G & C & mIOU & G & C & mIOU\\
			\hline
			Test RUGD (RUGD) &  82.93 &  67.74 &  58.09 &  \textbf{83.75} &  69.71 &  59.39 &  83.30 &  68.92 &  58.63 \\
			Test RUGD (MULT) &  82.61 &  68.76 &  57.28 &  83.32 &  \textbf{70.04} &  \textbf{59.70} &  83.23 &  69.94 &  58.83 \\
			\hline
			Test DeepScene (RUGD) &  70.03 &  59.72 &  43.58 &  67.62 &  56.55 &  40.83 &  66.75 &  60.40 &  41.84 \\
			Test DeepScene (MULT) &  93.03 &  85.27 &  76.21 &  93.47 &  85.66 &  78.11 &  \textbf{93.68} &  \textbf{86.08} &  \textbf{78.91} \\
			\hline
			Test YCOR (RUGD) &  74.95 &  50.43 &  41.55 &  76.41 &  53.07 &  43.89 &  76.24 &  53.58 &  44.22 \\
			Test YCOR (MULT) &  85.79 &  63.95 &  55.86 &  \textbf{86.93} &  68.70 &  \textbf{60.55} &  86.70 &  \textbf{69.56} &  60.47 \\
            \hline
			Test RELLIS (RUGD) &  76.58 &  39.88 &  30.93 &  76.24 &  37.14 &  29.30 &  76.65 &  39.04 &  30.79 \\
			Test RELLIS (MULT) &  89.46 &  68.16 &  57.02 &  89.85 &  \textbf{69.46} &  \textbf{57.89} &  \textbf{90.57} &  67.13 &  57.74 \\
			\hline
		\end{tabular}
	}
	\begin{tablenotes}
	\item\textbf{} 
    \item\textbf{Table Notes: } (RUGD) represents models only trained on the RUGD \\dataset while (MULT) represents models trained on multiple datasets.
    \end{tablenotes}
    \vspace{-6pt}
\end{table}

We focus more on analysis of test data performance since validation data is used to select models. We observe that combining different datasets for training improves model performance. We also observe that training with epistemic uncertainty improves model performance by a significant margin. Figure~\ref{fig:uqvis} shows some example predictions and entropy of model predictions. We discover that training with epistemic uncertainty leads to better calibrated model because entropy is higher when the model generates incorrect predictions. Since different datasets have different label classes and styles, there is additional noise introduced by combining datasets, we conjecture that treating this noise as epistemic uncertainty during training and inference contributes to a better model performance. From the Table \ref{tbl:testperformance}, it is challenging to determine whether MC Dropout or Masksembles is a better model in terms of quantitative performance. After consideration, we select Masksembles for field testing since it is more deterministic compared with MC Dropout.

\subsection{Platform setup}
\begin{figure}[h]
    \centering
    \includegraphics[width=0.35\textwidth]{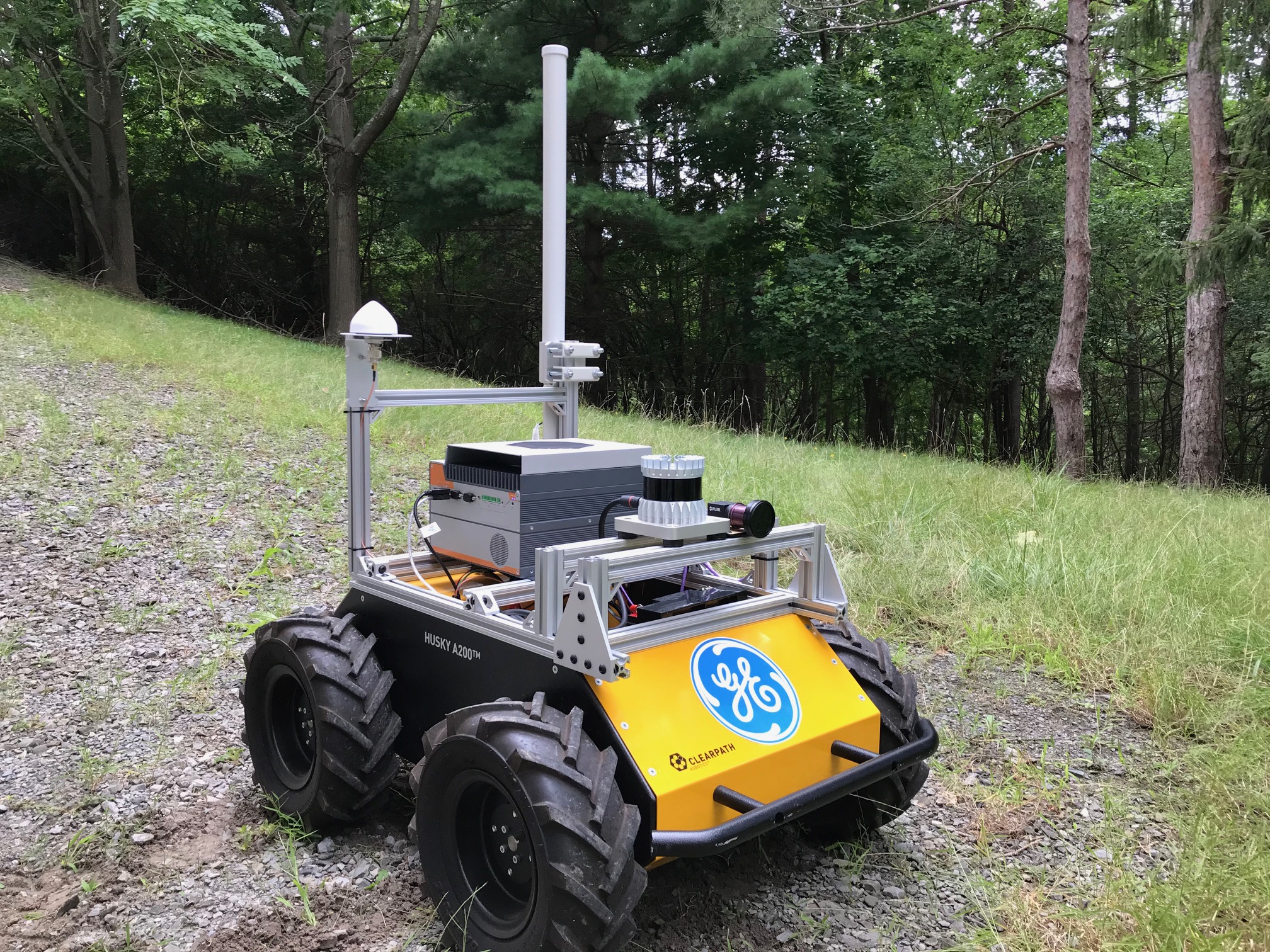}
    \caption{The research platform used for field tests.}
    \label{fig:husky_pic}
\end{figure}

The trained model was deployed on a Clearpath Husky robot for performing evaluation in off-road environment. The platform was running a software stack that was developed by the Combat Capabilities Development Command~(DEVCOM) Army Research Laboratory~(ARL)~\cite{Gregory2016}. We augmented the stack's perception module with our model that was trained with epistemic uncertainty, and fused the model's output uncertainty information in costmap generation to affect the path planning module~\cite{tan2021risk}. The robot (Figure~\ref{fig:husky_pic}) has a maximum speed of $1 m/s$, and is equipped with the following compute and sensor payload: 
\begin{itemize}
        \item Lidar: 64 beam Ouster OS-1
        \item Camera: FLIR Blackfly-S BFS‐PGE‐16S2C‐CS
        \item IMU: LORD Microstrain 3DM-GX3-25
        \item Hardware Time Synchronization: Masterclock GMR 1000 providing PTP server to lidar and cameras, PPS signal to IMU
        \item Intel Xeon E-2176G (Coffee Lake) 3.7 Ghz Processor with Zotac Nvidia GeForce GTX 1660 GPU, 32 GB RAM, and 256 GB storage
\end{itemize}

\subsection{Field Testing Results}
\begin{figure}[h!]
	\begin{center}
		\includegraphics[width=0.99\columnwidth]{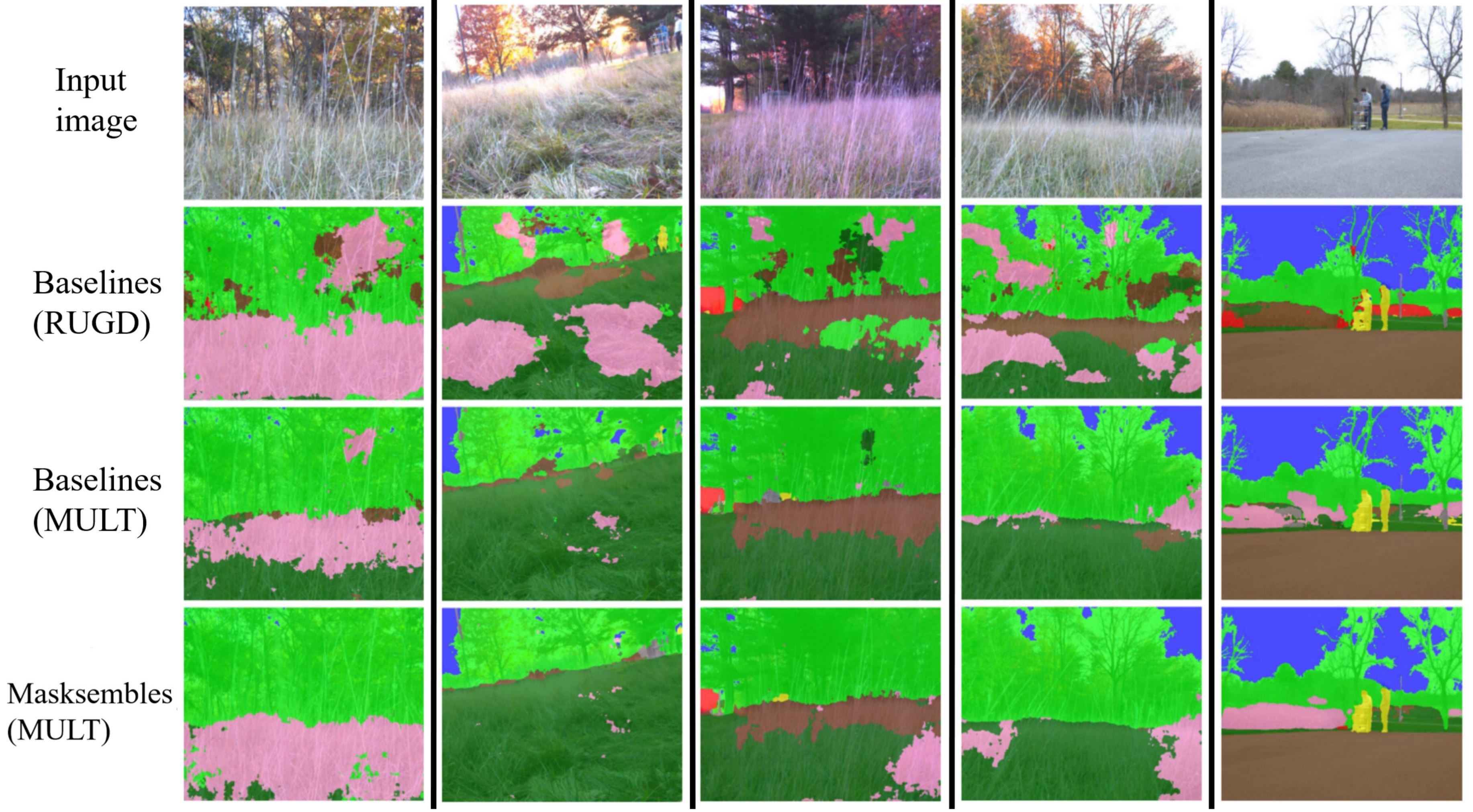}
	\end{center}
	\vspace{-10pt}
	\caption{Comparison of segmentation results for different models at our testing site. Each column contain an example image and its segmentation predictions from different models. First row is input images and second to forth rows are predictions from different models where (RUGD) represents models only trained on the RUGD dataset while (MULT) represents models trained on multiple datasets.}
	\label{fig:samplepredGRC}
	\vspace{-10pt}
\end{figure}

\begin{figure}[h!]
	\begin{center}
		\includegraphics[width=0.95\columnwidth]{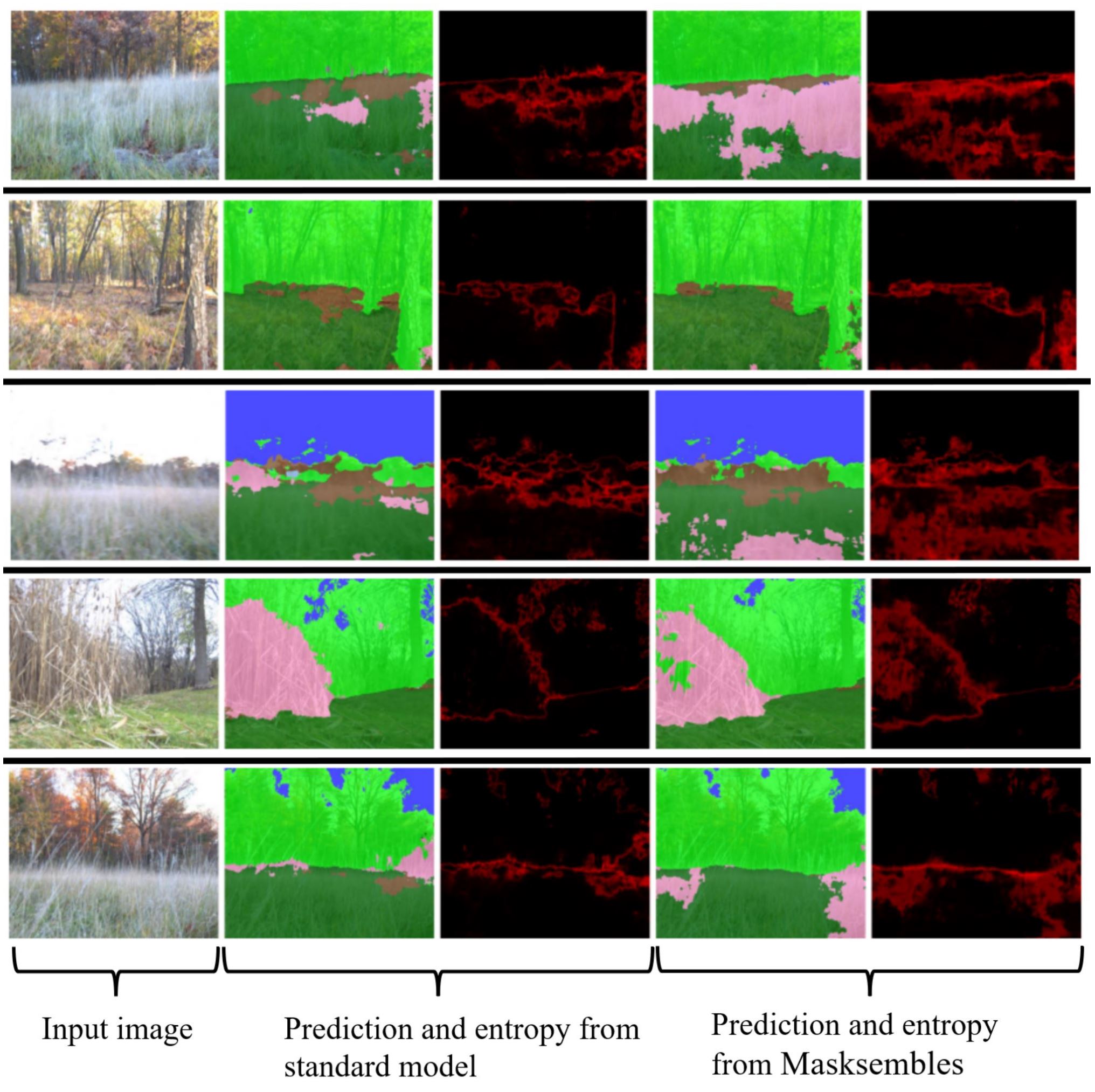}
	\end{center}
	\vspace{-10pt}
	\caption{Uncertainty estimation of images collected from our testing site. This is consider to be data with distribution drift and we observe Masksembles' predictions have higher entropy.}
	\label{fig:UQGRC}
	\vspace{-10pt}
\end{figure}

We performed various autonomous navigation experiments over multiple weeks in our local testing sites in upstate New York where environments are considered to be quite different with any datasets available during training. The experiments include traversing uneven terrains and slopes, across different vegetation and terrain types. The perception module using our resultant model was able to plan paths through these off-road condition and terrain types. Figure \ref{fig:samplepredGRC} shows some example segmentation results from the standard model only trained with the RUGD dataset, the standard model trained with multiple datasets and the Masksembles trained with multiple datasets. Figure \ref{fig:UQGRC} shows example predictions and entropy of predictions from our testing site, where the environments are considered to be different than any of the training datasets. Our results qualitatively indicate that training with multiple datasets and epistemic uncertainty will produce more reliable predictions.

\section{Conclusion and Future Work}
In this work, we investigate impact of training semantic segmentation models using multiple off-road datasets. We find that training models with a single dataset is not reliable for off-road autonomous navigation since most of public datasets are not collected in similar environments. Although, na\"\i vely combining multiple datasets can improve the model performance across those datasets, label noise negatively impacts the performance. The key insight from the presented work is that uncertainty-aware models perform better than traditional segmentation models when trained on multiple datasets. Our conjecture is that combining multiple datasets adds noise and inconsistencies that the uncertainty-aware models can effectively capture. Additionally, we found that if an uncertainty-aware model is tested against novel data, it is more likely to generate a higher entropy in predictions compared with a standard model. Our future work will further explore the impact of the uncertainty on other components in the autonomy stack such as the downstream navigation and planning routines that ingest the perception information. Ultimately, an improved characterization of the environment and uncertainty in perception should enable more robust navigation through unstructured off-road environments. 

\section*{ACKNOWLEDGMENT}
We gratefully acknowledge the Army Research Laboratory (ARL) Team: Andre Harrison, John Rogers, Long Quang, Maggie Wigness, and Eric Spero for their support, insights, and suggestions during the program. This work is also related to GE Humble AI initiative.

\bibliographystyle{unsrt}
\bibliography{main}

\begin{thebibliography}{10}

\bibitem{RUGD2019IROS}
Maggie Wigness, Sungmin Eum, John~G Rogers, David Han, and Heesung Kwon.
\newblock A {RUGD} dataset for autonomous navigation and visual perception in
  unstructured outdoor environments.
\newblock In {\em International Conference on Intelligent Robots and Systems
  (IROS)}, 2019.

\bibitem{tan2021risk}
Yewteck Tan, Nurali Virani, Brandon Good, Steven Gray, Mohammed Yousefhussien,
  Zhaoyuan Yang, Katelyn Angeliu, Nicholas Abate, and Shiraj Sen.
\newblock Risk-aware autonomous navigation.
\newblock In {\em Artificial Intelligence and Machine Learning for Multi-Domain
  Operations Applications III}, volume 11746, pages 335--348. SPIE, 2021.

\bibitem{cai2022risk}
Xiaoyi Cai, Michael Everett, Jonathan Fink, and Jonathan~P How.
\newblock Risk-aware off-road navigation via a learned speed distribution map.
\newblock {\em arXiv preprint arXiv:2203.13429}, 2022.

\bibitem{jiang2021rellis}
Peng Jiang, Philip Osteen, Maggie Wigness, and Srikanth Saripalli.
\newblock Rellis-3d dataset: Data, benchmarks and analysis.
\newblock In {\em 2021 IEEE international conference on robotics and automation
  (ICRA)}, pages 1110--1116. IEEE, 2021.

\bibitem{valada16iser}
Abhinav Valada, Gabriel Oliveira, Thomas Brox, and Wolfram Burgard.
\newblock Deep multispectral semantic scene understanding of forested
  environments using multimodal fusion.
\newblock In {\em International Symposium on Experimental Robotics (ISER)},
  2016.

\bibitem{maturana2018real}
Daniel Maturana, Po-Wei Chou, Masashi Uenoyama, and Sebastian Scherer.
\newblock Real-time semantic mapping for autonomous off-road navigation.
\newblock In {\em Field and Service Robotics}, pages 335--350. Springer, 2018.

\bibitem{SOOR2021}
Orighomisan Mayuku, Brian Surgenor, and Joshua Marshall.
\newblock Multi-resolution and multi-domain analysis of off-road datasets for
  autonomous driving.
\newblock pages 165--172, 05 2021.

\bibitem{ORFD2022}
Chen Min, Weizhong Jiang, Dawei Zhao, Jiaolong Xu, Liang Xiao, Yiming Nie, and
  Bin Dai.
\newblock {ORFD}: {A} dataset and benchmark for off-road freespace detection.
\newblock In {\em 2022 International Conference on Robotics and Automation
  (ICRA)}, pages 2532--2538, 2022.

\bibitem{Feng2021}
Di~Feng, Christian Haase-Sch\"{u}tz, Lars Rosenbaum, Heinz Hertlein, Claudius
  Gl\"{a}ser, Fabian Timm, Werner Wiesbeck, and Klaus Dietmayer.
\newblock Deep multi-modal object detection and semantic segmentation for
  autonomous driving: Datasets, methods, and challenges.
\newblock {\em Trans. Intell. Transport. Syst.}, 22(3):1341–1360, mar 2021.

\bibitem{Egocentric2022}
Galadrielle Humblot-Renaux, Letizia Marchegiani, Thomas~B. Moeslund, and Rikke
  Gade.
\newblock Navigation-oriented scene understanding for robotic autonomy:
  Learning to segment driveability in egocentric images.
\newblock {\em {IEEE} Robotics and Automation Letters}, 7(2):2913--2920, apr
  2022.

\bibitem{Cityscapes2016}
Marius Cordts, Mohamed Omran, Sebastian Ramos, Timo Rehfeld, Markus Enzweiler,
  Rodrigo Benenson, Uwe Franke, Stefan Roth, and Bernt Schiele.
\newblock The cityscapes dataset for semantic urban scene understanding.
\newblock In {\em Proceedings of the IEEE conference on computer vision and
  pattern recognition}, pages 3213--3223, 2016.

\bibitem{BDD2020}
Fisher Yu, Haofeng Chen, Xin Wang, Wenqi Xian, Yingying Chen, Fangchen Liu,
  Vashisht Madhavan, and Trevor Darrell.
\newblock {BDD100K}: A diverse driving dataset for heterogeneous multitask
  learning.
\newblock In {\em Proceedings of the IEEE/CVF Conference on Computer Vision and
  Pattern Recognition (CVPR)}, June 2020.

\bibitem{Mapillary2017}
Gerhard Neuhold, Tobias Ollmann, Samuel Rota~Bulo, and Peter Kontschieder.
\newblock The {Mapillary Vistas} dataset for semantic understanding of street
  scenes.
\newblock In {\em Proceedings of the IEEE International Conference on Computer
  Vision (ICCV)}, Oct 2017.

\bibitem{acdc2021}
Christos Sakaridis, Dengxin Dai, and Luc Van~Gool.
\newblock {ACDC}: {T}he adverse conditions dataset with correspondences for
  semantic driving scene understanding.
\newblock In {\em Proceedings of the IEEE/CVF International Conference on
  Computer Vision}, pages 10765--10775, 2021.

\bibitem{TAS2021}
Kai~A Metzger, Peter Mortimer, and Hans-Joachim Wuensche.
\newblock A fine-grained dataset and its efficient semantic segmentation for
  unstructured driving scenarios.
\newblock In {\em 2020 25th International Conference on Pattern Recognition
  (ICPR)}, pages 7892--7899. IEEE, 2021.

\bibitem{IDD2018}
Girish Varma, Anbumani Subramanian, Anoop Namboodiri, Manmohan Chandraker, and
  CV~Jawahar.
\newblock {IDD}: {A} dataset for exploring problems of autonomous navigation in
  unconstrained environments.
\newblock In {\em 2019 IEEE Winter Conference on Applications of Computer
  Vision (WACV)}, pages 1743--1751. IEEE, 2019.

\bibitem{Youngsaeng2021}
Youngsaeng Jin, David Han, and Hanseok Ko.
\newblock Memory-based semantic segmentation for off-road unstructured natural
  environments.
\newblock In {\em 2021 IEEE/RSJ International Conference on Intelligent Robots
  and Systems (IROS)}, pages 24--31, 2021.

\bibitem{Ojala2002}
T.~Ojala, M.~Pietikainen, and T.~Maenpaa.
\newblock Multiresolution gray-scale and rotation invariant texture
  classification with local binary patterns.
\newblock {\em IEEE Transactions on Pattern Analysis and Machine Intelligence},
  24(7):971--987, 2002.

\bibitem{kendall2017uncertainties}
Alex Kendall and Yarin Gal.
\newblock What uncertainties do we need in {Bayesian} deep learning for
  computer vision?
\newblock {\em Advances in neural information processing systems}, 30, 2017.

\bibitem{gal2016dropout}
Yarin Gal and Zoubin Ghahramani.
\newblock Dropout as a {Bayesian} approximation: Representing model uncertainty
  in deep learning.
\newblock In {\em international conference on machine learning}, pages
  1050--1059. PMLR, 2016.

\bibitem{lakshminarayanan2017simple}
Balaji Lakshminarayanan, Alexander Pritzel, and Charles Blundell.
\newblock Simple and scalable predictive uncertainty estimation using deep
  ensembles.
\newblock {\em Advances in neural information processing systems}, 30, 2017.

\bibitem{BMVC2017_57}
Alex Kendall, Vijay Badrinarayanan, and Roberto Cipolla.
\newblock Bayesian {SegNet}: Model uncertainty in deep convolutional
  encoder-decoder architectures for scene understanding.
\newblock In Gabriel~Brostow Tae-Kyun~Kim, Stefanos~Zafeiriou and Krystian
  Mikolajczyk, editors, {\em Proceedings of the British Machine Vision
  Conference (BMVC)}, pages 57.1--57.12. BMVA Press, September 2017.

\bibitem{durasov2021masksembles}
Nikita Durasov, Timur Bagautdinov, Pierre Baque, and Pascal Fua.
\newblock Masksembles for uncertainty estimation.
\newblock In {\em Proceedings of the IEEE/CVF Conference on Computer Vision and
  Pattern Recognition}, pages 13539--13548, 2021.

\bibitem{xiao2019}
Xuesu Xiao, Jan Dufek, and Robin Murphy.
\newblock Explicit-risk-aware path planning with reward maximization.
\newblock {\em arXiv preprint arXiv:1903.03187}, 2019.

\bibitem{pereira2013}
Arvind~A Pereira, Jonathan Binney, Geoffrey~A Hollinger, and Gaurav~S Sukhatme.
\newblock Risk-aware path planning for autonomous underwater vehicles using
  predictive ocean models.
\newblock {\em Journal of Field Robotics}, 30(5):741--762, 2013.

\bibitem{chung2019}
Jen Chung, Andrew~J Smith, Ryan Skeele, and Geoffrey~A Hollinger.
\newblock Risk-aware graph search with dynamic edge cost discovery.
\newblock {\em The International Journal of Robotics Research},
  38(2-3):182--195, 2019.

\bibitem{ronneberger2015u}
Olaf Ronneberger, Philipp Fischer, and Thomas Brox.
\newblock U-net: Convolutional networks for biomedical image segmentation.
\newblock In {\em International Conference on Medical image computing and
  computer-assisted intervention}, pages 234--241. Springer, 2015.

\bibitem{tan2019efficientnet}
Mingxing Tan and Quoc Le.
\newblock Efficientnet: Rethinking model scaling for convolutional neural
  networks.
\newblock In {\em International conference on machine learning}, pages
  6105--6114. PMLR, 2019.

\bibitem{Yakubovskiy2019}
Pavel Iakubovskii.
\newblock Segmentation models for {T}ensorflow and {K}eras.
\newblock https://github.com/qubvel/segmentation\_models, 2019.

\bibitem{lin2017focal}
Tsung-Yi Lin, Priya Goyal, Ross Girshick, Kaiming He, and Piotr Doll{\'a}r.
\newblock Focal loss for dense object detection.
\newblock In {\em Proceedings of the IEEE international conference on computer
  vision}, pages 2980--2988, 2017.

\bibitem{Gregory2016}
Jason Gregory, Jonathan Fink, Ethan Stump, Jeffrey Twigg, John Rogers, David
  Baran, Nicholas Fung, and Stuart Young.
\newblock {\em Application of Multi-Robot Systems to Disaster-Relief Scenarios
  with Limited Communication}, pages 639--653.
\newblock Springer International Publishing, 2016.

\end{thebibliography}

\addtolength{\textheight}{-12cm}   






\end{document}